# Sentiment Analysis of Lithuanian Online Reviews Using Large Language Models

Brigita Vileikytė[a], Mantas Lukoševičius[a] and Lukas Stankevičius[a]

*Kaunas University of Technology, Faculty of Informatics, Studentų St. 50, Lithuania*

**Abstract**
Sentiment analysis is a widely researched area within Natural Language Processing (NLP), attracting significant interest due to the advent of automated solutions. Despite this, the task remains challenging because of the inherent complexity of languages and the subjective nature of sentiments. It is even more challenging for less-studied and less-resourced languages such as Lithuanian. Our review of existing Lithuanian NLP research reveals that traditional machine learning methods and classification algorithms have limited effectiveness for the task. In this work, we address sentiment analysis of Lithuanian five-star-based online reviews from multiple domains that we collect and clean. We apply transformer models to this task for the first time, exploring the capabilities of pre-trained multilingual Large Language Models (LLMs), specifically focusing on fine-tuning BERT and T5 models. Given the inherent difficulty of the task, the fine-tuned models perform quite well, especially when the sentiments themselves are less ambiguous: 80.74% and 89.61% testing recognition accuracy of the most popular one- and five-star reviews respectively. They significantly outperform current commercial state-of-the-art general-purpose LLM GPT-4. We openly share our fine-tuned LLMs online.

**Keywords** [1]
Sentiment analysis, Opinion mining, Deep Learning, Lithuanian Online Reviews, Transformer model, BERT, T5, multilingual LLMs

## 1. Introduction

Nowadays, social media is often an integral part of people's lives, and for many, life is challenging. The growing number of internet users increases new online data resources and encourages growing online services. The Internet is changing fast to "read-write" mode [1]. With constantly growing consumer groups and online platforms, improving and maintaining excellent customer satisfaction levels is crucial. Since users are now more openly sharing their experiences, more and more people decide to buy products based on online reviews. More than 94% of people have confirmed that negative online reviews have persuaded them to avoid business [2]. Therefore, many companies want to analyze customer feedback to enhance their service and online presence. Automated solutions are in demand due to the challenge of content analysis. However, there is limited research on resources for languages like Lithuanian. Our paper focuses on sentiment analysis approaches for Lithuanian language-specific sentiment classification.

Text classification remains a cornerstone of Natural Language Processing (NLP) research. This broad field includes various tasks such as language identification, fraud detection, and categorization. Sentiment analysis, in particular, has become a highly studied area. The significance of opinion mining increased markedly following the influential research paper "*Thumbs Up?*" published by Bo Pang, Lillian Lee, and Shivakumar Vaithyanathan in 2002 [3]. After that, many scientists who studied this field applied various classification methods to increase accuracy in these tasks. The first study applied supervised classifier approaches like *Logistic Regression*, *Naïve Bayes Classification*, and *Support Vector Machines*. While with time, these methods combined in hybrid models started to improve their

---

performance, they still rely heavily on the features extracted from the text or matching hand-picked lexicons, that lack flexibility [4].

A significant breakthrough in NLP happened in 2014 after I. Sutskever, O. Vinyals, and Q. Le published the "*Sequence to Sequence Learning with Neural Networks*" article. The authors proposed a model that combined multiple layers of Recurrent Neural Networks (RNNs) in the encoder and decoder architecture and described a new approach to gradient clipping [5]. The same year a trainable neural attention mechanism first introduced in [6] that allowed the model to pick (attend to) RNN states corresponding to the words in an arbitrary order. Three years later, "*Attention Is All You Need*", by A. Vaswani and members of Google, introduced the transformer architecture [7], a deep learning model, that relied heavily and expanded upon on the neural attention and discarded of the RNN layers altogether. This shifted the predominant structure of deep learning NLP models. The introduced multi-headed self-attention mechanism enables the model to determine the significance of each word in the input text, facilitating the recognition and creation of complex contextual relationships within the text. This capability is crucial in developing prominent *Large Language Models* (LLMs) such as BERT and GPT-4 [8]. LLMs now solve numerous complex tasks for text generation, classification, etc., and the latest approaches aim to build on pre-trained models with relevant context, substantially increasing the accuracy of classifications.

Lithuanian sentiment analysis is not widely studied, and more research needs to be done. This is related to the size of the population and the available data on the Internet. When performing NLP case studies, English, Chinese, and German are the most considered languages. Lithuanian usually does not reach the top 10 or even 20 languages when considering the case studies made [9].

This work aims to use and fine-tune LLMs for Lithuanian-language-specific sentiment classification. We use two different transformer architectures for the classification task to set performance baselines and compare the quality of the models. This paper offers a succinct overview of sentiment analysis, focusing on the unique challenges we face when applying it to less common languages, discussed in Section 2. In Section 3, we delve into related work in machine learning and explore the latest methodologies in NLP classification tasks. In Section 4, we detail the methodology, models, and metrics we used for evaluation. Finally, in Section 5, we present the results of our experiments, and in Section 6, we suggest avenues for further improvements.

## 2. Sentiment analysis task and challenges

According to Webster's dictionary, "*sentiment*" encompasses multiple synonyms, including "*opinion*", "*emotion*", and "*view*". When conducting sentiment analysis, it is important to discern between subjective sentiments and those conveyed objectively. Opinions do not always express sentiments, and subjective statements with an opinion do not necessarily express emotions. Text with emotion might not have any associated opinion or sentiment [10]. Therefore, to structure the task better, "*sentiment analysis*" is commonly defined as a case study for automatic analysis of text evaluation and tracking of expressed judgments and feelings in text [11].

When working on a sentiment classification task, we first identify the type of sentiment classification we focus on. In literature, we often specify three classes: *document*, *sentence*, and *entity* level. When analyzing online reviews, we refer to *document-level* sentiment analysis. This type of analysis classifies a single review, where the input source is the whole text [12]. When classifying longer word sequences, it becomes crucial to identify the essential features and analyze words and their context. Often, supervised learning methods are applied to this task. The difficulties come from overcoming various linguistic and syntax challenges, possible sarcasm, co-references, and adjusting predictions for contrasting domains [13]. Therefore, when trying to accommodate already explored NLP methods in Lithuanian reviews, besides common issues, we also need to be aware of the data variations and linguistic complexity of the Lithuanian language.

## 3. Related work

Working with NLP tasks involves a combination of steps, and depending on the field of study and requirements, each step faces various challenges. This section reviews the main strategies used for text classification. We briefly describe the main steps required for data preprocessing and preparation, and the approaches taken for sentiment analysis classification using LLMs. The last subsection is dedicated to relevant work in NLP tasks with the Lithuanian language.

## 3.1. Text preprocessing and representation

A. TOKENIZATION

Tokenization splits text into smaller text units called *tokens,* and it is a crucial step in NLP preprocessing since LLMs cannot work with words directly. Tokenizing input text allows LLMs to handle more complex, prominent languages. Tokenization provides a structured way to break down text into manageable pieces for the model [14]. Modern tokenization approaches employed in transformer model architectures generally obviate the need for traditional text preprocessing that was typically applied to datasets when using conventional classification algorithms. Historically, standard preprocessing techniques included the removal of stop-words, stemming, and lemmatization. However, contemporary large language models are believed to handle various word forms effectively through advanced tokenization processes. Recent studies have demonstrated that retaining all unmodified input sequences yields better classification outcomes when employing models such as DistilBERT, BERT, and XLNet [15]. Different tokenization algorithms cater to specific computational needs and vocabulary requirements. In our research, we employ tokenization algorithms that align with the pre-training of these models.

*WordPiece*. The *WordPiece* model is a subword-based tokenization approach. This data-driven algorithm guarantees a deterministic segmentation for any possible sequence of characters [16]. Subword-based tokenizers first split the text by word segments. Therefore, whitespace information is neglected, and the tokenization process is irreversible [17]. *WordPiece* is an iterative algorithm; it starts with a combination of small vocabulary and special tokens. It then considers the frequency of words and their combinations and iteratively merges the most frequent parts. This helps to capture morphological information and generalize across different word forms. It can be computationally expensive as it picks the best pair at each iteration; despite this, it is quite popular because of great results [18]. *WordPiece* tokenization algorithm is used in the BERT model.

*SentencePiece*. *SentencePiece* is a language-independent subword tokenizer. Its lossless tokenization design allows information to be fully reversed to the input text before the tokenization. This is done by escaping whitespaces with a meta symbol and first tokenizing the input text into an arbitrary subword sequence [19]. This tokenizer implements the *Subword Regularization* algorithm [20]. *SentencePiece* adopts a $O(N log(N))$ computational cost algorithm where *N* is the length of the input sequence. This is significantly faster than most common tokenization algorithms based on byte pair encoding segmentation. Many popular LLM models like XLNet, T5, and LLaMa use the *SentencePiece* algorithm.

B. WORD EMBEDDINGS

Modern NLP systems heavily rely on pre-trained word embeddings. This approach offered significant improvements over embeddings learned by the models themselves. Word embeddings are dense feature vector representations in a specific-dimensional space. They are usually discovered by unsupervised algorithms when trained with a large amounts of text [21]. These methods are used in transformers as the first layer. We need to have our data tokenized to use the pre-trained word embedding algorithms. Therefore, the *WordPiece* embedding method is used in the BERT model, and in T5, the *SentencePiece* embedding is used.

## 3.2. LLMs for text classification

As mentioned in the Introduction section, the attention mechanism allowed models to weigh different words' relevance in the given input sequence differently. Models could finally capture complex relationships between words. This approach was initially applied as a translation mechanism [6]. After this article, further improvements involved removing LSTM parts, eliminating bottlenecks of encoder vectors, and enabling sequential processing. This work laid the foundation for models like

BERT and GPT-4. Transfer learning, where a model is pre-trained on a large amount of data before being fine-tuned on a downstream task, has shown excellent results and has become a powerful technique in NLP. After introducing the BERT model, it took almost five years for transformer models to be widely applied in daily use.

A. *BERT*

BERT stands for "*Bidirectional Encoder Representations from Transformers*", an encoder-only Transformer [22]. By the name, this model uses bidirectional context for pertaining. This characteristic allows the model to learn the context of a word based on the entire input context (left and right to the word in a sequence). It was first introduced by the Google team in 2018, and since then, several model variations have been applied to solving NLP tasks [22]. The model's ability to learn bidirectional context is used during training when the model learns to predict masked words from the context in the sequence. The BERT model can be applied to various language tasks like classification, question answering, and entity recognition. Numerous model size variations and complexities exist depending on the NLP task and resources [23].

B. *T5*

T5 stands for "*Text-to-Text transfer Transformer*", an encoder-decoder model. The basic T5 model treats every text processing problem as a "*text-to-text*" problem [24]. This model architecture takes text as input and produces new text as output. This approach is inspired by previous unifying frameworks for NLP tasks, including casting all text problems as question-answering [24]. The main unsupervised training objective is training the model to predict sentinel tokens previously purposely dropped out of the text. This general-purpose model is widely used in various NLP tasks when converting them to text-to-text problems [25].

## 3.3. Related work on the Lithuanian Language

Lithuanian sentiment analysis is not widely studied, and little research has been gathered. Currently, there are no monolingual LLMs pre-trained specifically in the Lithuanian language. However, the language is usually included in the multilingual model pre-training phases for the most popular models, like BERT. As of March 2024, the *HuggingFace* [2] platform has over 420 transformer models pre-trained with a subset of data for the Lithuanian language. Even with these resources, there are not many studies on sentiment analysis.

In recent years, sentiment analysis research on the Lithuanian language has increased due to advancements in NLP. Most studies can be found in the Electronic Academic Library of Lithuania, focusing on sentiment analysis in defined domain datasets. When comparing classical classification methods, accuracy for Lithuanian datasets usually does not exceed 80%. In a 2019 case study on Lithuanian Internet comments, Naive Bayes and Support Vector Machine algorithms outperformed LSTM and Convolutional neural network approaches, with accuracies of 0.735 and 0.724, respectively. The dataset consisted of 3 positive, negative, and neutral review categories [26]. Additionally, some interesting articles tried to adapt hybrid learning approaches to Lithuanian reviews. However, even with two or three classes to categorize, these results have not outperformed classical methods compared to the English language [27]. More research has been done on statistical and traditional neural network results. Still, with a lack of quality datasets and better research, we are unaware of significant sentiment classification improvements made in this topic.

With no monolingual Lithuanian LLM models publicly available, this language has limited baselines for sentiment classification. Nevertheless, some sentiment classification benchmarks are made considering other niche language results.

## 4. Data and Methodology

This section describes the collected data and explores the sentiment classification approach. For this work, we tried to keep the data as intact as possible during the cleaning and preparation steps before tokenizing it.

---

[2] https://huggingface.co/

## 4.1. The dataset

The data was collected from three different open online review sources: *pigu.lt*, *atsiliepimai.lt*, and *google.com/maps*. The main requirements for the sources were to target the Lithuanian language and to have a *5-point* or *5-star* rating system. We selected data sources with already defined categories since sentiment classification can be difficult even for humans and no one can indicate the sentiment better then the author of the comment him-/herself. The scraped user responses include a variety of subjects like restaurant and shop reviews as well as films and theme parks. The original dataset had 132,261 five-star-max reviews. We aimed to predict the marked ratings based on the review text alone for supervised sentiment analysis. Using the review rating as a sentiment label is common in sentiment analysis.

Even though we collected reviews targeting Lithuanians and sources with mainly registered users, there was still a lot of spam, language inconsistencies, and disarray in the data. To improve the quality of the data, the following main steps have been taken:
1. *Data anonymization*. We anonymized all data entries, leaving only reviews and their ratings.
2. *Data cleaning*. We removed all empty comments that had no alphabetic characters or contained only emoji symbols. This step discarded over 3,000 entries.
3. *Language selection*. Scraped data, even when targeting Lithuanian comments, consisted of other languages. Firstly, the *Python "langid"* library was used to assign languages for each review. Any languages other than Lithuanian were mapped to a separate list. This dataset consisted of over 45,000 entries. Not all categories from the separated list were identified correctly (*mostly short LT phrases were misclassified*). Therefore, we handled some cases by hand and used *GPT-4 API* prompting to identify review languages automatically. We translated straightforward reviews that were no longer than five words. Finally, we discarded over 10,000 reviews in this step.
4. *Data size selection*. We dropped all entries that had been longer than 450 words to remove extrema from data and have consistent, comparable datasets among various models.

The final data size consisted of 123,604 entries. Figure 1 shows the distribution of reviews based on their categories in the prepared dataset. This type of distribution is commonly seen in practice. People tend to write reviews when they are delighted or disappointed with the products [28].

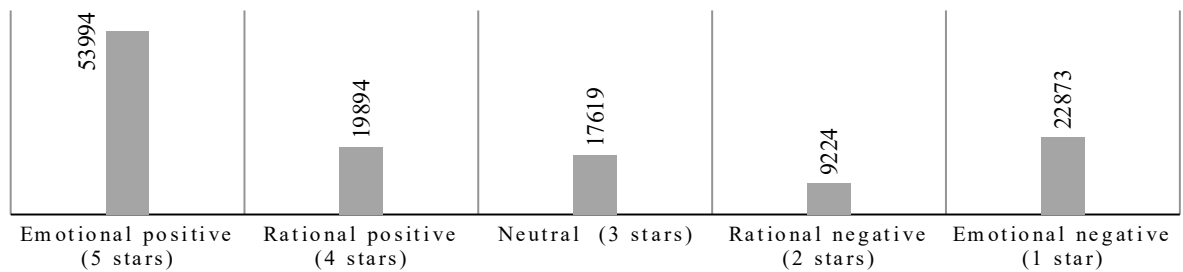

**Figure 1**: Sentiment category distribution in the dataset

99.3% of reviews are shorter than 150 words. The most significant outliers in text length variations are in emotionally negative comments. A box plot by category identified the most extensive interquartile range in emotionally negative comments. The five-star category is the most symmetric, but scraped reviews generally have a more skewed distribution based on word count.

Lastly, we created two train, evaluation, and test datasets. The cleaned dataset shows a repeated category distribution for the first group, while the second dataset focuses on more defined categories. Since Lithuanian synonym and augmentation tools are not readily available, we down-sampled our positive reviews, removed duplicates, and tried to discard the most prominent similar reviews. Table 1 shows the final test, train, and evaluation data set sizes and distributions by category.

**Table 1**
Dataset size and category distribution for different dataset samples

| Parameters Dataset nr. | Dataset size | | | Category distribution, % | | | | |
|---|---|---|---|---|---|---|---|---|
| | Train | Evaluation | Test | 5 | 4 | 3 | 2 | 1 |
| 1st | 84050 | 24721 | 14833 | 43,6 | 16,1 | 14,2 | 7,4 | 18,5 |
| 2nd | 55489 | 18497 | 18497 | 24,7 | 21,5 | 19,1 | 9,9 | 24,7 |

## 4.2. Methods used

We worked in the *Paperspace workspace*[3] to train our LLM models using an A6000[4] instance.

A. *BERT*

For experimentation, we analyzed a *DistilBERT-type* model. A distilled version of BERT models is designed to retain 97% accuracy while being 40% smaller and 60% faster [29]. With limited computational and data resources, using this subtype of BERT models is a commonly acquired practice. The model we chose for finetuning was "*distilbert-base-multilingual-cased*"[5]. When fine-tuning the model, we worked on a multilingual model trained on a Wikipedia dataset of 104 languages, including Lithuanian. The model has six layers, 768 dimensions, and 12 heads, totaling 134M parameters.

We initialize the DistilBERT model with a classification head. The top-level modules for the models are *distilbert*, *pre_classifier*, *classifier*, and *dropout*. During experiments, we noticed that fine-tuning the whole model and adjusting all the layers tended to overfit our small dataset quickly. We had chosen considerably high values of 0.3 and 0.2 for the sub-layer and attention dropout.

B. *T5*

For our experiments, we also worked with the ByT5 model. ByT5 is based on the mT5 model, which was trained on a large set of unlabeled multilingual text data. It has various model sizes, and the one we used is considered "*Small*". To improve our results with a limited training data set, we use the "*ByT5-Lithuanian-gec-100h*"[6] model that has been additionally trained on Lithuanian text from Lithuanian news articles [30]. This model was created during work towards Lithuanian grammar correction and had been trained on Lithuanian text for about 100 hours.

We adapted a text-to-text model for text classification. We present the task to the model as a text-generating task, and our sentiment labels are the expected model predictions. When analyzing generated results, we decode produced token IDs from the model. This is possible since the T5 uses a lossless tokenization algorithm. We then map the outputs to the expected numbered labels and calculate result metrics.

C. *Evaluation.*

For the final evaluation, we used:

- *Accuracy*. This metric measures the proportion of correctly classified entries in the total number of predictions made. It is straightforward to interpret but disregards class balances and the costs of other errors.
- *F1-score*. F1-score for a certain class identified the harmonic mean of its precision and recall. This allows us to more precisely evaluate the overall quality of a classifier's predictions. The score values range from 0 to 1, with 1 being the best score.

According to the confusion matrix, classification results can be staggered into four cases: *True Positive* (TP), *False Positive* (FP), *True Negative* (TN), and *False Negative* (FN). Then, the F1-score for a single class can be calculated as follows [31]:

$$Precision = \frac{TP}{FP+TP}, \quad (1)$$

$$Recall = \frac{TP}{FN+TP}, \quad (2)$$

$$F_1 = \frac{2*Precision*Recall}{Precision+Recall}. \quad (3)$$

To compute the global *F1-Score*, we should compute global precision and recall scores from the sum of TP, FP, TP, and TN across all classes. We then use these values to calculate the global F1 Score as their harmonic mean, a micro average mean.

---

[3] https://www.paperspace.com/notebooks
[4] A6000 GPU machine has 48GB of GPU, 8vCOUs, 45GB of CPU RAM
[5] https://huggingface.co/distilbert/distilbert-base-multilingual-cased
[6] https://huggingface.co/LukasStankevicius/ByT5-Lithuanian-gec-100h

## 5. Results

All experiments done during this work have been applied to the two Lithuanian datasets. The final experiment results are presented in *Table 2*.

When experimenting with the BERT model, we mainly used between 10 and 20 epochs since the epoch training time and computational resources were smaller and more viable to handle. When experimenting with T5, we used at most ten epochs on each run, considering that the model quickly overfitted the training data. Moreover, each epoch took significantly more computation and time resources than the BERT model.

**Table 2**
Experiment results for DistilBERT and ByT5 models

| Model | Dataset | Number of epochs | Evaluation results | | Test results | |
|---|---|---|---|---|---|---|
| | | | Accuracy | F1 score | Accuracy | F1 score |
| distilBERT | 1$^{st}$ | 12 | 68,43% | 68,02% | 67,41% | 67,51% |
| | 2$^{nd}$ | 10 | 62,01% | 60,23% | 62,31% | 61,7% |
| ByT5 | 1$^{st}$ | 5 | 65,31% | 61,25% | 63,26% | 60,32% |
| | 2$^{nd}$ | 6 | 60,27% | 57,81% | 59,95% | 57,66% |

When fine-tuning T5 and DistilBERT models for sentiment analysis, the experiments highlighted the importance of dataset composition. Both models demonstrated improved performance on datasets with more imbalanced distribution of classes. However, a tendency for quick overfitting to the training data was observed in both cases. Even with a larger dataset, the validation loss of the distilBERT model started to increase significantly from the 10$^{th}$ to the 12$^{th}$ epoch. ByT5 exhibited early-stage overfitting around the 4$^{th}$ or 5$^{th}$ epoch. Notably, the DistilBERT model, when trained on the entire dataset, exhibited significant overgeneralization, predominantly predicting positive sentiments. This suggests a need for careful dataset selection and the implementation of strategies to mitigate overfitting during the training process. We can see that the best results have been achieved by fine-tuning the distilBERT model with 1$^{st}$ dataset. We achieved a 67.51% F1 score with the distilBERT model on the test dataset.

**Table 3**
Confusion matrix results for test dataset of DistilBERT model

| True label \ Prediction | Emotionality negative | Rationally negative | Neutral | Rationally positive | Emotionally positive |
|---|---|---|---|---|---|
| Emotionality negative | 2135 (80.74%) | 248 (9.38%) | 197 (7.45%) | 82 (3.10%) | 83 (3.14%) |
| Rationally negative | 362 (26.32%) | 402 (29.20%) | 232 (16.85%) | 71 (5.15%) | 40 (2.91%) |
| Neutral | 237 (12.76%) | 217 (11.69%) | 984 (53.00%) | 396 (21.31%) | 280 (15.08%) |
| Rationally positive | 48 (2.63%) | 32 (1.75%) | 299 (16.41%) | 1030 (56.51%) | 978 (53.60%) |
| Emotionally positive | 71 (1.14%) | 25 (0.40%) | 149 (2.37%) | 590 (9.39%) | 5645 (89.61%) |

Table 3 displays the DistilBERT model test data predictions using a confusion matrix. The table presents the model's classification outcomes by comparing its predictions with the sentiment values. It reveals that the model best predicts emotionally negative (☆) and emotionally positive (☆☆☆☆☆) sentiments. However, misclassifications are more prevalent in the neutral (☆☆☆) and rationally negative (☆☆) categories, with the highest number of incorrect predictions.

When the sentiment categories are reduced from the five to three (negative, neutral, positive), the overall accuracy improves, especially in correctly identifying negative and positive comments. The distilBERT model, trained on the initial dataset, accurately predicts negative comments (75.79% accuracy) and identifies positive comments (91.01% accuracy) when evaluating the same fine-tuned model with 3-class sentiment analysis.

Finally, we used a well-known, production-ready GPT-4 model to evaluate its sentiment classification accuracy on the 2$^{nd}$ dataset. The model achieved an accuracy of 55.18% and an F1 score of 0.5012, indicating difficulties distinguishing specific sentiment categories, especially rationally negative comments. Significant inaccuracies were found in classifying neutral and rationally positive sentiments. Despite simplifying the three categories, the accuracy improved to 73.41% with an F1 score of 0.7013, yet the performance remained below average compared with fine-tuned DistilBERT and ByT5 models.

## 6. Discussion

When working on LLM models, it is crucial to have pre-trained models that are relevant to the task context. Currently, there are no known monolingual Lithuanian language LLM models openly available. We understand that creating and pre-training a monolingual model is a computationally expensive and time-consuming task. Moreover, we could still need more training data with limited internet resources on the Lithuanian language compared to the more popular ones like English or Chinese. This could help the LLMs to generalize specific language in more detail and extract greater language context features. We believe this could be further explored to reach more prominent results.

## 7. Conclusions

In this work, we analyze the current trends in NLP classification tasks. We collected a new multi-topic Lithuanian dataset for customer reviews with a five-star rating system. The dataset was scraped from various external sources to simulate real-world scenarios more closely. We used this dataset to experiment with two types of LLM. We analyzed multilingual DistilBERT and ByT5 model capabilities for sentiment classification tasks on the Lithuanian language dataset.

After summarizing our experiments, we found that the fine-tuned distilBERT model reached better results on the unseen test dataset. The free open-source model outperformed the GPT-4, highlighting that even current commercial state-of-the-art general-purpose LLMs language models like GPT-4 can be lacking compared to specialized models without specific fine-tuning.

The classification results show that the five-labels sentiment classification is a complex task even for LLMs. Our dataset represents real-world situations and has some hard-to-recognize patterns, even for humans, when no additional context is provided, especially for the intermediate ratings. The textual and the stars-given parts of the review compliment, not substitute each other, therefore the sentiment expressed is not always identical in both. For example, a user may give a four-star rating and a quite negative textual review explaining why a single star has been subtracted. Given the knowledge that the data has limitations and obstacles for the subjectiveness of users and their experiences, our fine-tuned models showed substantially good results.

Further work could be done to improve dataset quality, size, and model parameters, potentially leading to even better outcomes.

We share our fine-tuned models at https://huggingface.co/brivil1/lithuanian-sentiment-analysis-DistilBERT and https://huggingface.co/brivil1/lithuanian-sentiment-analysis-ByT5.